\documentclass[10pt,twocolumn,letterpaper]{article}

\usepackage{cvpr}              

\usepackage[dvipsnames]{xcolor}

\usepackage{graphicx}
\usepackage{amsmath}
\usepackage{amssymb}
\usepackage{booktabs}

\usepackage{multirow}
\usepackage{subcaption}
\usepackage{url}
\usepackage{adjustbox}
\usepackage{boxedminipage}
\usepackage{listings}

\definecolor{cvprblue}{rgb}{0.21,0.49,0.74}
\usepackage[pagebackref,breaklinks,colorlinks,citecolor=cvprblue]{hyperref}

\def\paperID{2828} 
\def\confName{CVPR}
\def\confYear{2024}

\title{Depth Insight - Contribution of Different Features to \\ Indoor Single-image Depth Estimation}

\author{Yihong Wu\textsuperscript{$1$}, Yuwen Heng\textsuperscript{$1,2$}, Mahesan Niranjan\textsuperscript{$1$}, Hansung Kim\textsuperscript{$1$}\\
\textsuperscript{$1$} School of Electronics and Computer Science, University of Southampton\\
\textsuperscript{$2$} Baidu Inc.\\
}

\begin{document}
\maketitle
\begin{abstract}
Depth estimation from a single image is a challenging problem in computer vision because binocular disparity or motion information is absent. Whereas impressive performances have been reported in this area recently using end-to-end trained deep neural architectures, as to what cues in the images that are being exploited by these black box systems is hard to know. To this end, in this work, we quantify the relative contributions of the known cues of depth in a monocular depth estimation setting using an indoor scene data set. Our work uses feature extraction techniques to relate the single features of shape, texture, colour and saturation, taken in isolation, to predict depth. We find that the shape of objects extracted by edge detection substantially contributes more than others in the indoor setting considered, while the other features also have contributions in varying degrees. These insights will help optimise depth estimation models, boosting their accuracy and robustness. They promise to broaden the practical applications of vision-based depth estimation. The project code is attached to the supplementary material and will be published on GitHub.
\end{abstract}    
\section{Introduction}
\label{sec:intro}

Depth estimation, predicting the distance from an object’s surface to the camera, is a key task in the field of computer vision. It plays a crucial role in many applications, such as 3D reconstruction \cite{alawadh2022room}, autonomous driving \cite{janai2020computer,wang2019pseudo}, virtual reality (VR) \cite{dickson2021benchmarking}, augmented reality (AR) \cite{lee2011depth}, etc. The goal of monocular depth estimation based on deep learning is to infer the depth value of each pixel by analysing the scene information in a single image.

Due to the ill-posed problem of monocular depth estimation, there is a fundamental need to move towards a scene understanding of objects in images so that various characteristics of objects can cue depth information. Recent work on monocular depth estimation using end-to-end trained deep neural network models shows that such cues are collectively learnable and satisfactory depth estimation can be achieved \cite{eigen2014depth,alhashim2018high,bhat2021adabins,wu2023depth}. However, the black-box nature of such models prohibits the understanding of what cues are exploited in monocular depth estimation. The mechanism behind monocular depth estimation based on 2D images in neural networks is still not clearly explained, and the extent to which these models can approximate the human capability of monocular depth perception remains uncertain.

Building on this gap in understanding, and inspired by causality analysis \cite{liu2022structural}, we aimed to investigate the factors that influence depth estimation. This work will pave the way for the development of versatile models applicable to a broader spectrum of depth estimation tasks, moving beyond reliance solely on data-driven approaches. Research has shown that monocular depth cues in 2D images include phenomena such as blurring, shading and brightness \cite{swain1997integration}. This paper investigated and analysed the factors that influence machine-based monocular depth estimation. To provide a comprehensive understanding, we investigated the roles of the factors relevant to object recognition \cite{ge2022contributions}, such as colour and texture, in the context of monocular depth estimation. However, many factors are interrelated and cannot be independently segregated. In the context of scenarios where it is possible to directly and independently extract features from 2D images, we have considered colour, saturation, texture and shape, and each of them holds significant relevance in image processing, exerting varying degrees of impact on the overall outcome.


\noindent \textbf{Colour.}
Colour is recognised by the perception and interpretation of different wavelengths of light by the eye \cite{grzybowski2019color}. The visual information humans gather heavily relies on the presence of colour \cite{neitz2000molecular}. Colour helps humans recognise and remember objects faster \cite{gegenfurtner2000sensory}. 
Nevertheless, when defining colour, it is critical to recognise that RGB images do not only represent a singular colour but also include various elements in addition to colour, such as shape and texture. To isolate the pure colour information, we utilised a phase scrambling approach \cite{ge2022contributions}, which effectively separates the colour from these additional attributes.

\begin{figure}[tb]
  \centering
\begin{subfigure}[b]{0.48\linewidth}
    \centering
    \includegraphics[width=\linewidth]{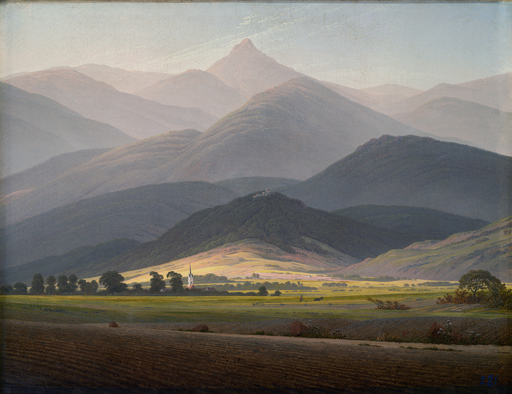}
    \caption{Nature Scene Image}
    \label{aerialPerspectiveNature}
  \end{subfigure}
  \hspace{-0.5em}
  \begin{subfigure}[b]{0.48\linewidth}
    \centering
    \includegraphics[width=\linewidth]{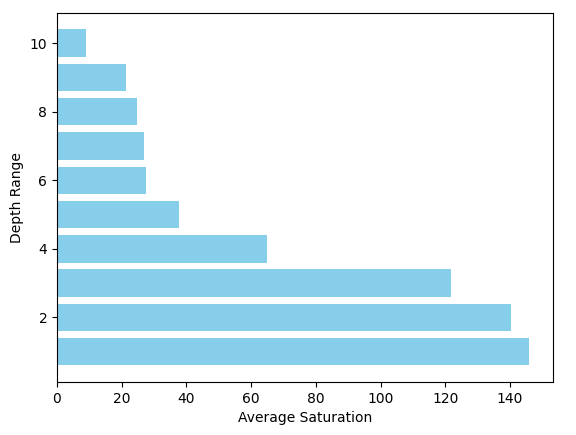}
    \caption{Saturation Distribution}
    \label{aerialPerspectiveSaturation}
  \end{subfigure}
    \caption{Saturation Analysis for a Nature Scene}
    \label{aerialPerspective}
\end{figure}

\noindent \textbf{Saturation.}
The second feature of interest is saturation. Saturation refers to the purity or intensity of a colour. For instance, high saturation indicates a more vivid and pure colour, while low saturation suggests a lighter or more desaturated colour with a hint of grey. Aerial perspective, within the domain of remote viewing, refers to the impact of the atmosphere on the visual depiction of an object. For instance, in Figure \ref{aerialPerspectiveNature}, a nature photograph is displayed. We evenly split images into ten rows, and the average saturation values have been calculated for each row, as depicted in Figure \ref{aerialPerspectiveSaturation}. As the object moves away from the camera, it can be observed that the saturation decreases. Building on this observation, saturation serves as a depth cue for outdoor single-image depth estimation. We aimed to investigate the utility of saturation as a depth cue in indoor scenes. 

\noindent \textbf{Texture.}
In computer vision, texture is defined by repetitive patterns with varying intensities present in an image\cite{tuceryan1993texture}. Prior research has found that textures are important when influencing a human's perception of distance \cite{rowland1999effects}, with specific regions in the brain having been found to be activated when exposed to varying textures \cite{puce1996differential}. Therefore, we also sought to independently extract the features pertaining to texture and assess their impact on depth estimation.

\noindent \textbf{Shape.}
A shape is generally considered to be a graphical representation of an object or its external borders, contours or external surfaces. Acquiring precise boundaries of objects in the 3D world based solely on 2D images is challenging. To simplify this process, we defined the shape feature as the edge graph, which corresponds to a greyscale map generated using an edge detection algorithm designed to preserve the object's boundaries. Edges are regarded as one of the primary cues essential for the human visual system \cite{farid2013edges}. Edge graphs usually represent geometric structures or boundaries between objects. For depth estimation tasks, the geometric features of an object are crucial to inferring its depth. The geometric structure aids depth estimation algorithms in capturing the shapes and relationships between objects \cite{jin2020geometric}, with edge maps providing supplementary geometric information. Edge detectors can analyse pixel gradients in different areas of the image, thereby assisting in the estimation of the relative distances between objects.
\section{Related Works}
\label{sec:related_work}

Interpretability within deep learning is attracting significant and growing interest. Interest in Convolutional Neural Networks (CNNs) and Visual Transformers has been rapidly increasing lately, particularly concerning their interpretability. A study revealed that CNN models trained on ImageNet exhibit a heightened sensitivity to texture information \cite{geirhos2018imagenet}. In addition, recent research has undertaken a comparison of the attributes between CNNs and Transformers across various layers \cite{raghu2021vision} by using Centred Kernel Alignment (CKA) \cite{cortes2012algorithms,kornblith2019similarity}. According to their claims, the transformer allows the early gathering of global information in contrast to CNNs. This results in a robust propagation of features from lower to higher layers in the network. Nevertheless, the primary focus of these enquiries remains centred on model analysis.

In the realm of human depth perception, substantial work has been conducted to investigate cues such as position in the image, texture density and focus blur \cite{gibson1950perception,cutting1995perceiving}. Existing works have demonstrated various methods for indoor single-image depth estimation that exhibit good performance \cite{eigen2014depth,bhat2021adabins}. Despite this, an analysis of their operations is still lacking. To the best of our knowledge, there has been no analysis of the contributions of different types of depth cues specifically for deep learning-based depth estimation in indoor single-image scenarios. In the two most relevant prior studies to our work, \cite{hu2019visualization} conducts attribution analysis to identify pixels that contribute most significantly to the final depth map. However, these methods can only offer insights into the low-level workings of CNNs. The analysis in \cite{dijk2019neural} was primarily confined to specific objects situated in outdoor environments, such as animals and vehicles on roadways. In contrast, in our study, we focused on colour, saturation, texture and shape, taking into account that our target application pertains to indoor scenes and requires the extraction of these cues from a single image.

Approaches for emulating the human capacity for gauging depth from an indoor scene still encounter gaps in knowledge. The objective of this paper is to unveil how neural networks extract depth-related information from a single indoor image to attain a more profound comprehension of the disparities between monocular visual depth estimation and the depth perception exhibited by humans.

Simultaneously, our work offers a foundational framework to facilitate subsequent investigations into assessing the interdependence among pertinent variables in the realm of depth estimation. A prior exploration delved into the causal interplay within 3D reconstruction, deconstructing elements like perspective and depth while also attempting to substantiate the interlinkages amid diverse variables \cite{liu2022structural}. Nonetheless, the model expounded upon in this enquiry operates on the assumption that the object is symmetric \cite{wu2020unsupervised}. Through an autoencoder mechanism, it internally dissects the input image into manifold components, as opposed to explicitly extracting a corresponding viewpoint, depth and related insights from the RGB image. In our study, we exclusively extracted various factors from RGB images while carefully investigating the significance of these factors within the context of depth estimation. Our work is set to further enable causal analyses in the field of depth estimation, paving the way for future advancements in comprehending the causality of depth estimation.
\section{Methodology}

\begin{figure}[tb]
  \centering
\begin{subfigure}[b]{0.4\linewidth}
    \centering
    \includegraphics[width=\linewidth]{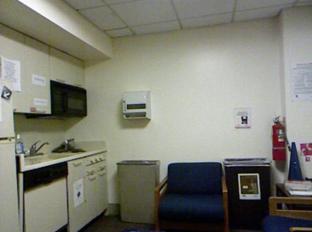}
    \caption{Original RGB}
    \label{fig:subfig1}
  \end{subfigure}
  \hspace{0em}
  \begin{subfigure}[b]{0.4\linewidth}
    \centering
    \includegraphics[width=\linewidth]{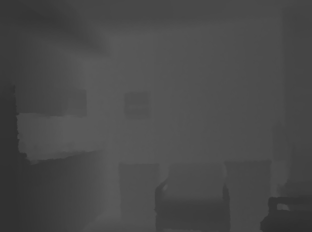}
    \caption{Corresponding GT Depth}
    \label{fig:subfig2}
  \end{subfigure}
    \caption{A Sample from NYU Dataset}
    \label{originalRGB}
\end{figure}

In this section, We consider four cues for monocular depth estimation. In order to compare the appearance of these four different features, we use the same sample in this section. Figure \ref{originalRGB} shows the original RGB image and its corresponding ground truth (GT) depth from the NYU data set \cite{silberman2012indoor}.

\subsection{Colour}

\begin{figure}[tb]
  \centering
  \begin{subfigure}[b]{\linewidth}
    \centering
    \includegraphics[width=\linewidth]{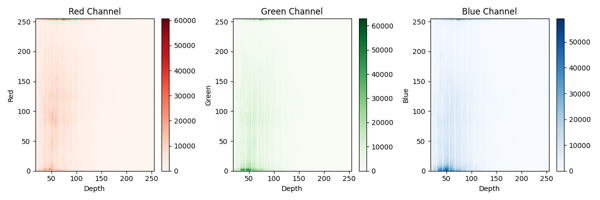}
  \end{subfigure}
    \caption{Heat map of the relationship between the distribution of original RGB three-channel values and the depth map. The horizontal axis represents the depth range, while the vertical axis corresponds to the pixel count of the R, G and B channels within the respective depth ranges. The colour bar values represent the pixel counts for three respective channels from 500 images randomly selected from the NYU data set.}
  \label{heatmaps_main}
\end{figure}

Figure \ref{heatmaps_main} illustrates the relationship between the distribution of original RGB three-channel values and the depth maps. The pixels on original RGB images are primarily concentrated between 0 and 100 in the corresponding grey-scale depth maps. The heat map reveals that the values of R, G and B pixels are similarly distributed across a specific depth range. This shows that the factors affecting depth are not significantly related to the distribution of pixels on the RGB channel. More details are shown in the Appendix.

\begin{figure}[tb]
  \centering
\begin{subfigure}[b]{0.4\linewidth}
    \centering
    \includegraphics[width=\linewidth]{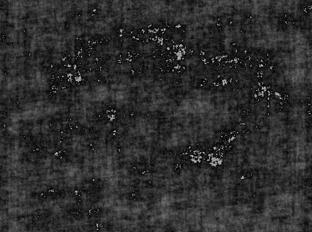}
    \label{discontinuity1}
  \end{subfigure}
  \hspace{0em}
  \begin{subfigure}[b]{0.4\linewidth}
    \centering
    \includegraphics[width=\linewidth]{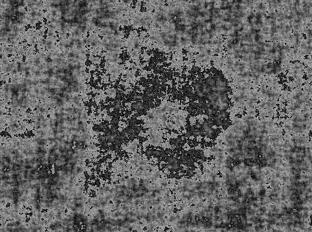}
    \label{discontinuity2}
  \end{subfigure}
    \caption{Phase Scrambled H Map and Corresponding Depth Map of Figure \ref{originalRGB}}
    \label{discontinuity}
\end{figure}

Hue from the hue, saturation and luminance value (HSV) colour space can be an expression of colour. However, hue values represent the projection of the RGB colour space onto a non-linear chroma angle \cite{szeliski2010computer}. If an output pixel value falls outside the valid range, it necessitates remapping to bring it within the specified range. The chroma angle represents a non-linear trajectory within a continuous, uninterrupted space. Here, starting at 0 degrees is the same as coming full circle to 360 degrees. However, when we apply this idea to an image, like with H maps, the smooth flow is interrupted, creating a series of separated points instead. Figure \ref{discontinuity} illustrates the images and corresponding depth maps resulting from the phase scrambling and remapping process applied to the H map from Figure \ref{originalRGB}, which are mapped back to specific intervals. Some discontinuous blocks can be observed in this figure. Therefore, We did not consider utilising the hue maps as the colour feature.

To examine the contribution of colour to depth, we 
performed phase scrambling \cite{ge2022contributions} on original RGB images and their corresponding depth maps to remove influences from shapes, textures and other geometric features. The resulting data set was labelled ``RGB Phase Scrambled''. Nevertheless, even after the phase scrambling, the outcome still retains the brightness information, making it not purely a colour feature. Subsequently, these outputs were converted to greyscale, effectively removing the colour information, and the resulting data set was labelled as ``Greyscale Phase Scrambled''. To illustrate the role of colour, a comparison of these two phase-scrambled features is conducted.

\begin{figure}[tb]
  \centering
\begin{subfigure}[b]{0.3\linewidth}
    \centering
    \includegraphics[width=\linewidth]{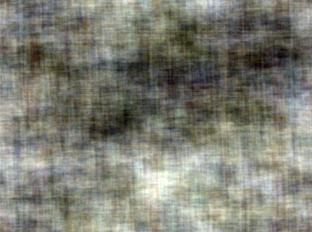}
    \caption{RGB Image Phase Scrambling}
    \label{rgb_phase_scrambling}
  \end{subfigure}
  \hspace{0em}
  \begin{subfigure}[b]{0.3\linewidth}
    \centering
    \includegraphics[width=\linewidth]{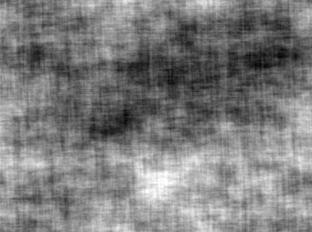}
    \caption{Gray-scale Image Phase Scrambling}
    \label{rgb_phase_scrambling_grayscale}
  \end{subfigure}
  \hspace{0em}
  \begin{subfigure}[b]{0.3\linewidth}
    \centering
    \includegraphics[width=\linewidth]{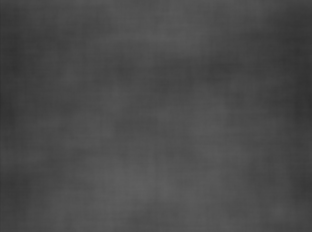}
    \caption{Depth Phase Scrambling}
    \label{rgb_phase_scrambling_depth}
  \end{subfigure}
    \caption{Phase Scrambling Results of Figure \ref{originalRGB}}
    \label{color}
\end{figure}

\subsection{Saturation}

\begin{figure}[tb]
    \centering
    \includegraphics[width=0.35\textwidth]{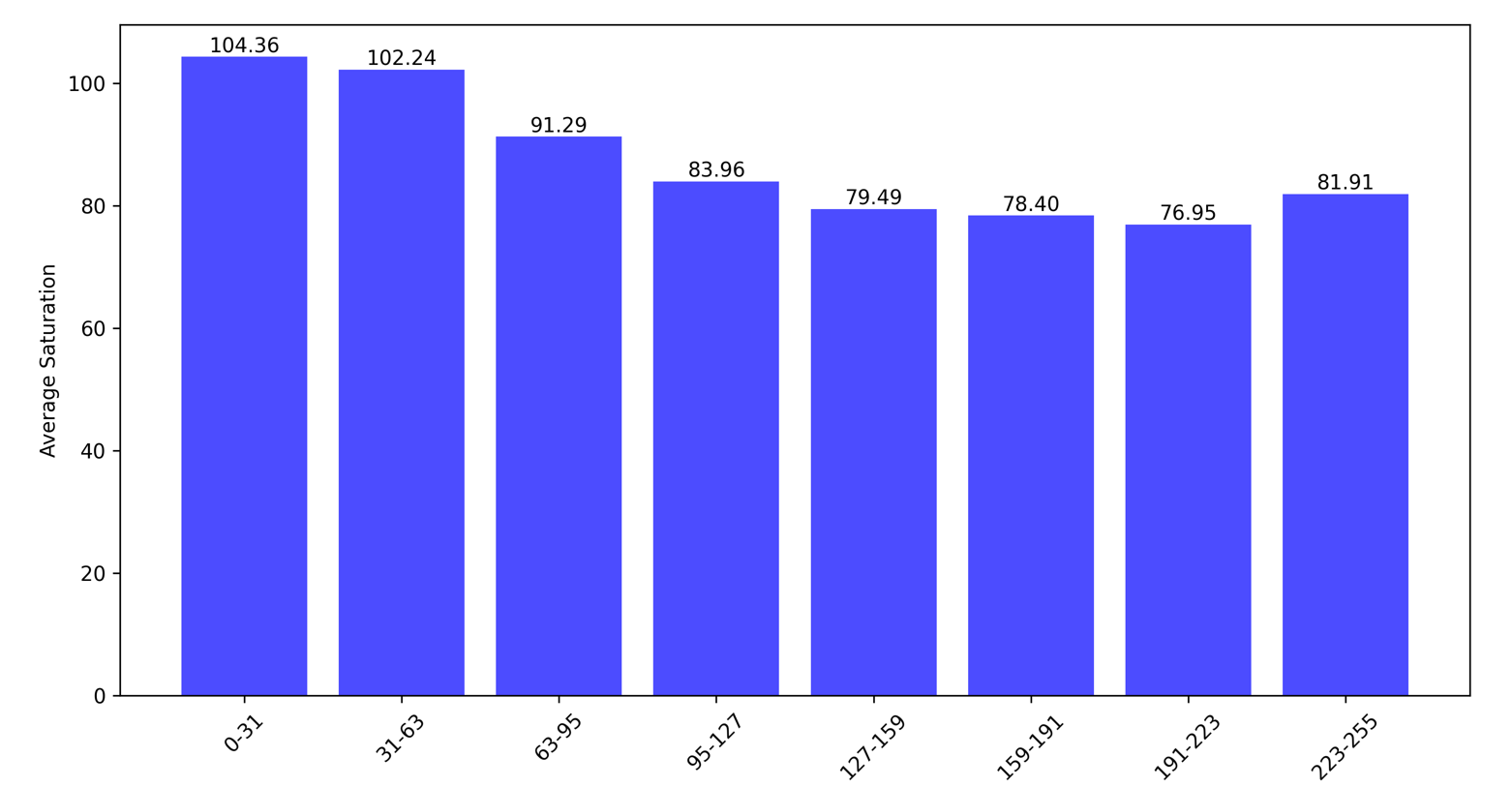}
    \caption{Average Saturation at Different Depth Intervals for Indoor Scenes (NYU data set)}
    \label{depth_range}
\end{figure}

We investigated whether saturation varies at different depths in indoor scenes. We partitioned this depth range 0-255 in the NYU data set into eight segments and then calculated the average saturation for each by converting RGB to HSV colour space and extracting the saturation values. Figure \ref{depth_range} shows the average saturation of the NYU data set in different depth ranges. Based on the observations, it appears that saturation may have less influence on the results for indoor scenes, different from the result for outdoor scenes shown in Figure \ref{aerialPerspective}. 

Nevertheless, we intended to further investigate the extent to which this subtle difference can affect depth estimation. In addition, as mentioned in the Introduction, human depth perception can be influenced by saturation. To assess the specific contribution of saturation, we extracted the saturation feature for experimentation independently.

To extract the features pertaining to saturation, we started by converting the RGB colour space to the HSV colour space and then extracting the saturation maps. Subsequently, these saturation maps are subjected to phase scrambling to eliminate features such as shape, texture and other visual characteristics.

\begin{equation}
\mathrm{V} \leftarrow \max (\mathrm{R}, \mathrm{G}, \mathrm{B}) \label{v}
\end{equation}

\begin{equation}
\mathrm{S} \leftarrow \begin{cases}\frac{\mathrm{V}-\min (\mathrm{R}, \mathrm{G}, \mathrm{B})}{\mathrm{V}} & \text { if } \mathrm{V} \neq 0 \\
0 & \text { otherwise }\end{cases}
\label{s}
\end{equation}

\begin{figure}[tb]
  \centering
\begin{subfigure}[b]{0.4\linewidth}
    \centering
    \includegraphics[width=\linewidth]{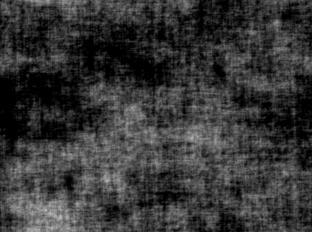}
    \caption{Saturation Feature}
    \label{saturation_feature}
  \end{subfigure}
  \hspace{0em}
  \begin{subfigure}[b]{0.4\linewidth}
    \centering
    \includegraphics[width=\linewidth]{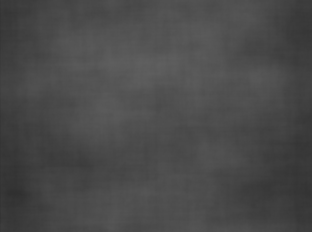}
    \caption{Corresponding Depth}
    \label{saturation_depth}
  \end{subfigure}
    \caption{Saturation with Phase Scrambling of Figure \ref{originalRGB}}
    \label{saturation}
\end{figure}

For each pixel, the V maps are obtained by taking the maximum value (Eq.\ref{v}) among the RGB channels. Subsequently, the saturation feature is obtained based on phase scrambling from S maps (Eq.\ref{s}). As depicted in Figure \ref{saturation}, Figure \ref{saturation_feature} illustrates the saturation feature, with Figure \ref{saturation_depth} displaying its corresponding depth map.

\subsection{Local Texture}

The preference for local textures over global textures stems from the fact that the extraction of global textures includes the consideration of additional factors, including shape and other features. To mitigate the influence of other factors and preserve the texture, the images were segmented into patches and shuffled, thus eliminating global information such as shapes, since this information introduces more features than just textures.

\subsection{Shape}

The boundaries of an object define its precise outline, marking the separation between the object and its immediate environment. Edge maps are generated through the analysis of gradient variations in image pixel values and identify changes in these values. Although edge maps do not always faithfully represent real object boundaries, when dealing with a single 2D image, they offer an efficient means of simulating object shapes. This feature has been defined as `shape' for the subsequent experiment.

\begin{figure}[tb]
  \centering
\begin{subfigure}[b]{0.3\linewidth}
    \centering
    \includegraphics[width=\linewidth]{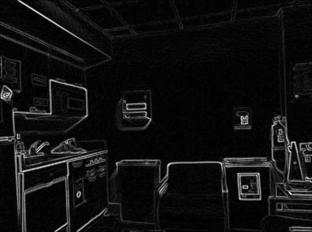}
    \caption{Shape with Sobel}
    \label{sobel}
  \end{subfigure}
  \hspace{-0.2em}
  \centering
\begin{subfigure}[b]{0.3\linewidth}
    \centering
    \includegraphics[width=\linewidth]{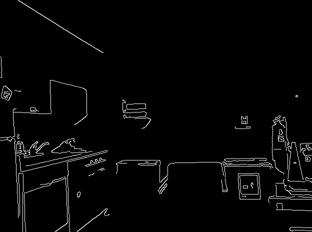}
    \caption{Shape with Canny}
    \label{canny}
  \end{subfigure}
  \hspace{-0.2em}
  \begin{subfigure}[b]{0.3\linewidth}
    \centering
    \includegraphics[width=\linewidth]{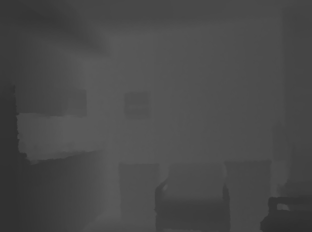}
    \caption{Depth}
    \label{shapeDepth}
  \end{subfigure}
    \caption{Shape Maps from Figure \ref{originalRGB} and Corresponding Ground Truth Depth}
    \label{shape}
\end{figure}

As shown in Figure \ref{shape}, we utilised the Canny operator instead of the Sobel operator because the latter will find the gradient in the x and y directions, reflecting the differential changes of pixels \cite{szeliski2010computer}. Therefore, not only the shape feature is included when using the Sobel operator, but some texture information may also be introduced.

\section{Experiments}

As mentioned in the Introduction, we considered four factors that may contribute to depth estimation: colour, saturation, local texture and shape. All of these features were trained using the baseline model, and the obtained results were analysed.

\subsection{Data}

We used the NYU dataset \cite{silberman2012indoor}, which serves as a widely employed dataset in computer vision, particularly for depth estimation research. Comprising images from diverse indoor scenes, it encompasses a variety of objects and furniture. The size of the NYU dataset enhances the representativeness of model training and evaluation. It is derived from 464 scenes in three cities. The resolution of the images is $640 \times 480$. 10\% of the data is split as the testing set.

\subsection{Model} \label{Model}

The UNet architecture is preferred for deep learning-based depth estimation due to its comprehensive design, adept at gathering context and integrating features across different scales \cite{bhat2021adabins, wu2023depth, alhashim2018high, eigen2014depth}. This preference is rooted in UNet's feature pyramid structure and efficient reuse of features, enhancing depth estimation by capturing diverse scale information while preserving detail. Our experiments demonstrate that employing ResNet50 as the backbone is sufficient for model convergence on our dataset. Subsequently, we utilised the U-Net network with ResNet50 as the backbone in the following experiment.

\subsection{Evaluation Metrics} \label{evaluation_metrics}

We utilised six metrics commonly used in the field of depth estimation, which include three accuracy metrics and three error metrics. The accuracy metrics are distinguished by thresholds at $1.25$, $1.25^2$ and $1.25^3$, each reflecting different levels of tolerance for deviation from the true values. Higher values of these accuracy metrics indicate better model performance. For error metrics, the absolute relative error ($rel$) quantifies the average deviation of predicted values from the actual values. The root mean squared error ($rmse$) can amplify the effect of outliers by taking the square root of the average of the squared deviations from the ground truth, and the logarithmic error ($log_{10}$) metric mitigates the impact of outliers by applying a logarithmic scale to the error values. Lower values of these error metrics signify superior model performance.

\subsection{Experiments and Analysis}

\begin{table*}[tb]
  \centering
  \caption{Depth Estimation Performance with Different Inputs}
  \begin{adjustbox}{width=0.95\textwidth,center}
  \begin{tabular}{ccccccc}
    \toprule
    Features & $a_{1}\uparrow$ & $a_{2}\uparrow$ & $a_{3}\uparrow$ & $log_{10}\downarrow$ & $rel\downarrow$ & $rmse\downarrow$ \\
    \midrule
    Original RGB Images & $98.13\pm0.0013$ & $99.64\pm0.0003$ & $99.9\pm0.0001$ & $0.0176\pm0.0001$ & $0.0413\pm0.0008$ & $0.0174\pm0.0002$ \\
    RGB Phase Scrambled & $43.5\pm0.0067$ & $72.64\pm0.0068$ & $87.47\pm0.0042$ & $0.1498\pm0.0021$ & $0.4754\pm0.017$ & $0.113\pm0.0019$ \\
    Grayscale Phase Scrambled & $36.13\pm0.0316$ & $64.09\pm0.041$ & $81.65\pm0.033$ & $0.1769\pm0.0143$ & $0.5627\pm0.0585$ & $0.1364\pm0.0149$ \\
    Saturation & $36.9\pm0.0094$ & $65.35\pm0.0113$ & $82.86\pm0.0065$ & $0.1718\pm0.0026$ & $0.5321\pm0.0208$ & $0.1296\pm0.0015$ \\
    Local Texture & $49.95\pm0.0286$ & $77.88\pm0.0228$ & $90.83\pm0.0114$ & $0.1276\pm0.0068$ & $0.3187\pm0.0215$ & $0.1065\pm0.0047$ \\
    Shape & $96.46\pm0.0003$ & $99.12\pm0.0002$ & $99.71\pm0.0002$ & $0.0235\pm0.0001$ & $0.0556\pm0.0004$ & $0.0224\pm0.0001$ \\
    \bottomrule
  \end{tabular}%
  \end{adjustbox}
  \label{allResults}%
\end{table*}%

Table \ref{allResults} presents the performance of depth estimation using different input features, evaluated by several metrics (details shown in Sec.\ref{evaluation_metrics}). Original RGB images performed the best with high accuracy ($a_1$, $a_2$, $a_3$) and low error ($log_{10}$, $rel$, $rmse$). Phase-scrambled RGB and greyscale images, along with saturation inputs, showed significantly worse performance, with greyscale being the least accurate. Inputs of local texture had moderate accuracy and error rates, while shape features performed close to the original RGB images.

\subsubsection{Colour}

\begin{figure}[tb]
  \centering
\begin{subfigure}[b]{0.3\linewidth}
    \centering
    \includegraphics[width=\linewidth]{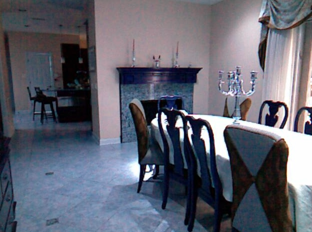}
    \caption{RGB Image}
    \label{colorOriginal}
  \end{subfigure}
  \hspace{0em}
  \begin{subfigure}[b]{0.3\linewidth}
    \centering
    \includegraphics[width=\linewidth]{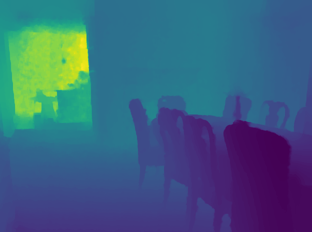}
    \caption{GT Depth}
    \label{colorDepth}
  \end{subfigure}
  \hspace{0em}
  \begin{subfigure}[b]{0.3\linewidth}
    \centering
    \includegraphics[width=\linewidth]{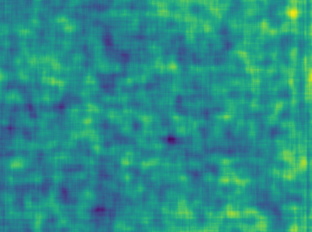}
    \caption{Estimated Depth}
    \label{colorOutput}
  \end{subfigure}
    \caption{Depth Estimation with a Colour Feature Input. The left and middle images are the original RGB image and the corresponding ground truth depth map, respectively. The image on the right depicts the estimated depth map, which is the result of the model's output after inverse phase scrambling, employing the colour feature as the input.}
    \label{restoredColour}
\end{figure}

To evaluate the contribution of the colour feature, we trained the model with phase-scrambled RGB images. Figure \ref{restoredColour} displays the original RGB image, ground truth depth map and the estimated depth map, the latter of which has been reconstructed from the scrambled image using a pre-stored random matrix. As aligned to the low accuracy indicated in Table \ref{allResults}, it is hard to recognise the original scene structure from the estimated depth. 

\begin{figure}[tb]
  \centering
  \begin{subfigure}[b]{0.35\linewidth}
    \centering
    \includegraphics[width=\linewidth]{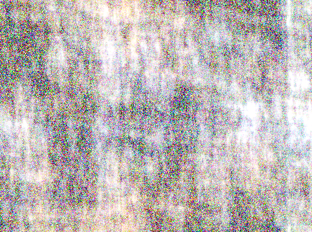}
  \end{subfigure}
  \hspace{0em}
  \begin{subfigure}[b]{0.35\linewidth}
    \centering
    \includegraphics[width=\linewidth]{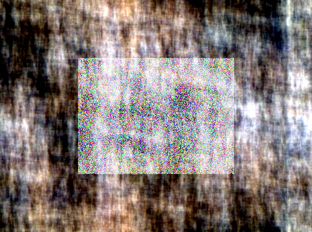}
  \end{subfigure}
\begin{subfigure}[b]{0.35\linewidth}
    \centering
    \includegraphics[width=\linewidth]{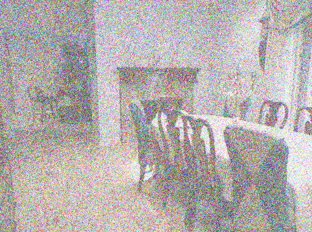}
    \caption{Noise for Whole Image}
    \label{noisedWhole}
  \end{subfigure}
  \hspace{0em}
  \begin{subfigure}[b]{0.35\linewidth}
    \centering
    \includegraphics[width=\linewidth]{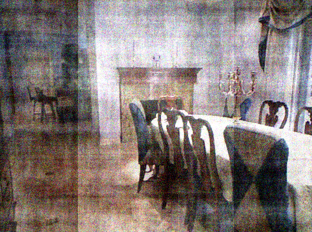}
    \caption{Noise in Central Region}
    \label{noisedCentral}
  \end{subfigure}
    \caption{Noised with Phase Scrambled Images. Figure \ref{noisedWhole} illustrates the outcome of applying Gaussian noise to the entire image and subsequently restoring it, while Figure \ref{noisedCentral} depicts the results of adding noise and restoring only the central area, where both the length and width are half of the original image's dimensions.}
    \label{noised}
\end{figure}

To simulate scenarios where the model output differs from the ground truth, we added Gaussian noise (mean = 0, std = 25) to the phase scrambled image. Figure 11 shows examples of our phase scrambled image with added Gaussian noise and their corresponding reconstructions. Figure \ref{noised} shows the outcomes of introducing Gaussian noise to the phase-scrambled image, followed by its restoration using the pre-stored random matrix. As we can see, despite the introduction of noise through phase scrambling, this noise does not affect the shape and position of objects in the recovered images. The performance in Figure \ref{colorOutput} can be attributed to the poor performance of the model when provided with colour phase-scrambled input.

Furthermore, by comparing the respective performances of ``RGB Phase Scrambled'' and ``Grayscale Phase Scrambled'' inputs as shown in Table \ref{allResults}, it can be observed that, after excluding the contribution of brightness, colour has a limited impact on depth estimation.

\subsubsection{Saturation}

\begin{figure}[tb]
  \centering
\begin{subfigure}[b]{0.3\linewidth}
    \centering
    \includegraphics[width=\linewidth]{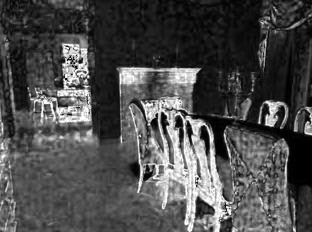}
    \caption{Saturation Map}
    \label{saturationOriginal}
  \end{subfigure}
  \hspace{0em}
  \begin{subfigure}[b]{0.3\linewidth}
    \centering
    \includegraphics[width=\linewidth]{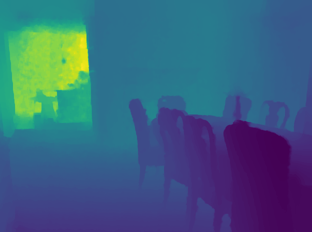}
    \caption{GT Depth}
    \label{saturationDepth}
  \end{subfigure}
  \hspace{0em}
  \begin{subfigure}[b]{0.3\linewidth}
    \centering
    \includegraphics[width=\linewidth]{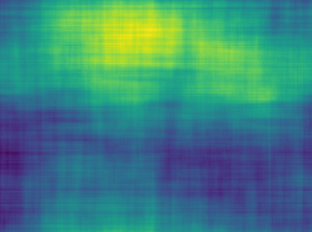}
    \caption{Estimated Depth}
    \label{saturationOutput}
  \end{subfigure}
    \caption{Depth Estimation with a Saturation Feature Input. The left and middle images are the original RGB image and the corresponding ground truth depth map, respectively. The image on the right depicts the estimated depth map, which is the result of the model's output after inverse phase scrambling, employing a saturation map as the input.}
    \label{restoredSaturation}
\end{figure}

We trained the baseline model with saturation maps as the input and evaluated the contribution of the saturation feature. Figure \ref{restoredSaturation} illustrates the saturation map, corresponding ground truth depth and the output from the restoration process. Similarly, due to the poor performance, the restored output lacks discernible features such as object contours.

Table \ref{allResults} shows that the $a_{1}$ is about 37\%. Although saturation contributed to estimating the depth of the indoor scene, its contribution was minor. Saturation exhibits lower accuracies compared to other features except for greyscale phase scrambled input. Furthermore, error metrics substantiate this observation. The $rel$ stands at 0.904, while the root mean square error ($rmse$) is shown as 0.1196. Therefore, using saturation as a measure for depth estimation clearly introduces a significant error compared to the true depth values. Despite its poor performance, saturation still plays a role in assessing depth in indoor scenes. This highlights that saturation can provide some depth cues in certain contexts, although it comes with a higher error margin.

\subsubsection{Local Texture}

\begin{figure}[tb]
  \centering
\begin{subfigure}[b]{0.3\linewidth}
    \centering
    \includegraphics[width=\linewidth]{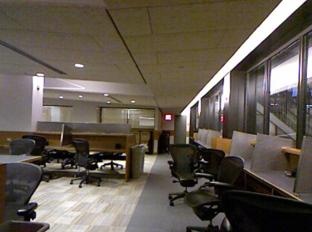}
    \caption{Original Image}
    \label{patchOriginal}
  \end{subfigure}
  \hspace{0em}
  \begin{subfigure}[b]{0.3\linewidth}
    \centering
    \includegraphics[width=\linewidth]{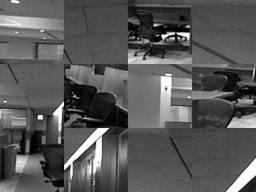}
    \caption{Patch Size 128}
    \label{patch128}
  \end{subfigure}
  \hspace{0em}
  \begin{subfigure}[b]{0.3\linewidth}
    \centering
    \includegraphics[width=\linewidth]{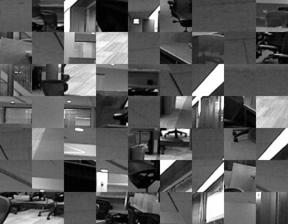}
    \caption{Patch Size 64}
    \label{patch64}
  \end{subfigure}
  \hspace{0em}
  \begin{subfigure}[b]{0.3\linewidth}
    \centering
    \includegraphics[width=\linewidth]{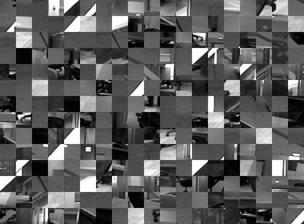}
    \caption{Patch Size 32}
    \label{patch32}
  \end{subfigure}
  \hspace{0em}
  \begin{subfigure}[b]{0.3\linewidth}
    \centering
    \includegraphics[width=\linewidth]{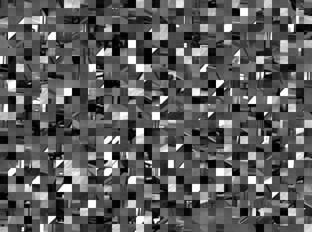}
    \caption{Patch Size 16}
    \label{patch16}
  \end{subfigure}
  \hspace{0em}
  \begin{subfigure}[b]{0.3\linewidth}
    \centering
    \includegraphics[width=\linewidth]{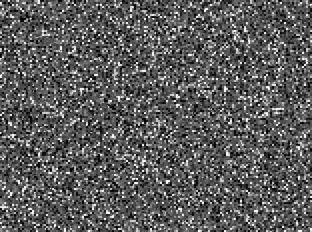}
    \caption{Patch Size 4}
    \label{patch4}
  \end{subfigure}
    \caption{Local Texture with Different Patch Sizes of a Random Sample}
    \label{texturePatches}
\end{figure} 

Variations in the field of view and resolution will impact the size of the patch used to extract local textures. The optimal patch size should align with the specific data set and scene scale employed. Figure \ref{texturePatches} illustrates the paths with varying patch dimensions. As shown in Figure \ref{patch128}, when we use a large patch size of 128, the texture itself is not isolated because the shape and context information still present in the patches. As the patch size is decreased, the shape of the objects in the image becomes less apparent and, therefore, the textures present in the image are increasingly segregated. As demonstrated in Figure \ref{patch16}, for the data set we used, the $16\times16$ patch size is particularly well-suited for texture extraction while minimising the influence of other features (e.g. shape). This size is large enough to restrict shape details but not so small as to be impractical, unlike the $4\times4$ patch depicted in Figure \ref{patch4}.

\begin{table*}[tb]
  \centering
  \caption{Performance with Different Patch Sizes}
  \begin{adjustbox}{width=0.85\textwidth,center}
  \begin{tabular}{ccccccc}
    \toprule
    Size & $a_{1}\uparrow$ & $a_{2}\uparrow$ & $a_{3}\uparrow$ & $log_{10}\downarrow$ & $rel\downarrow$ & $rmse\downarrow$ \\
    \midrule
    4 & $40.97\pm0.0037$ & $69.6\pm0.003$ & $85.99\pm0.0018$ & $0.1523\pm0.0008$ & $0.3812\pm0.0026$ & $0.1239\pm0.0007$ \\
    16 & $49.95\pm0.0286$ & $77.88\pm0.0228$ & $90.83\pm0.0114$ & $0.1276\pm0.0068$ & $0.3187\pm0.0215$ & $0.1065\pm0.0047$ \\
    32 & $53.27\pm0.042$ & $80.64\pm0.0253$ & $92.46\pm0.0104$ & $0.1185\pm0.0087$ & $0.3012\pm0.0266$ & $0.1009\pm0.0094$ \\
    64 & $74.22\pm0.0166$ & $92.48\pm0.0122$ & $97.51\pm0.0057$ & $0.0747\pm0.0039$ & $0.1885\pm0.0199$ & $0.0629\pm0.002$ \\
    128 & $93.12\pm0.0066$ & $98.47\pm0.0019$ & $99.5\pm0.0007$ & $0.0358\pm0.0016$ & $0.0863\pm0.0039$ & $0.0338\pm0.0015$ \\
    \bottomrule
  \end{tabular}%
  \end{adjustbox}
  \label{patches}%
\end{table*}%

Table \ref{patches} shows the performance of texture inputs (shuffle patches) in different patch sizes. As can be seen in the figure, the accuracy rate gradually increases with the increase of patch sizes in height. This is because a larger patch contains more information besides the texture, such as the shape of the object.

\begin{figure}[tb]
  \centering
\begin{subfigure}[b]{0.3\linewidth}
    \centering
    \includegraphics[width=\linewidth]{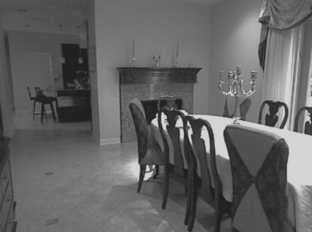}
    \caption{Gray-scale Image}
    \label{restoredTextureOriginal}
  \end{subfigure}
  \hspace{0em}
  \begin{subfigure}[b]{0.3\linewidth}
    \centering
    \includegraphics[width=\linewidth]{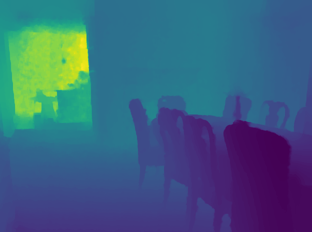}
    \caption{GT Depth}
    \label{restoredTextureGT}
  \end{subfigure}
  \hspace{0em}
  \begin{subfigure}[b]{0.3\linewidth}
    \centering
    \includegraphics[width=\linewidth]{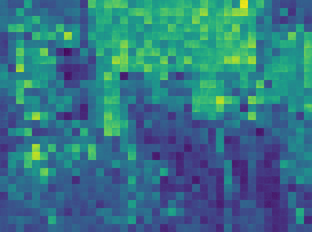}
    \caption{Estimated Depth}
    \label{restoredTextureOutput}
  \end{subfigure}
    \caption{Depth Estimation with a Local Texture Input with Patch Size $16\times16$. The left and middle images are the original RGB image and the corresponding ground truth depth map, respectively. The image on the right depicts the estimated depth map, which is the result of the model's output after inverse shuffle by using a pre-stored random matrix, employing a local texture feature as input.}
    \label{restoredTexture}
\end{figure}

A sample of the original greyscale image, along with the corresponding depth map and the estimated depth map, is displayed in Figure \ref{restoredTexture}, demonstrating the results of the model trained with local texture inputs. To focus on local textures during training, Figure \ref{restoredTextureOriginal} and Figure \ref{restoredTextureGT} are split into $16\times16$ patches and these patches are shuffled using the same random matrix to eliminate global scene information, such as object shapes. As shown in Figure \ref{restoredTextureOutput}, the estimated depth map only provides a coarse approximation of the scene's depth, distinguishing between nearer and farther areas but failing to capture the precise depth details.

The local texture appears as a minor factor in depth estimation, yielding an $a_{1}$ accuracy of a mere 50\% in Table \ref{allResults}. Error metrics also show this trend although they are slightly better than the colour and saturation features. This issue happens because the position changes to the patches weaken their connections, thereby making it harder for the model to understand objects and context. This proposition finds corroboration in the robust performance observed upon deploying the shape feature as the input data source in Sec \ref{shape_experiment_section}.

\subsubsection{Shape} \label{shape_experiment_section}

\begin{figure}[tb]
  \centering
\begin{subfigure}[b]{0.3\linewidth}
    \centering
    \includegraphics[width=\linewidth]{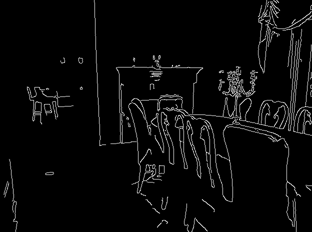}
    \caption{Shape Map}
    \label{ShapeOriginal}
  \end{subfigure}
  \hspace{0em}
  \begin{subfigure}[b]{0.3\linewidth}
    \centering
    \includegraphics[width=\linewidth]{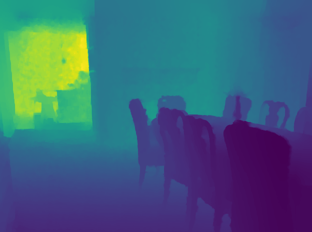}
    \caption{GT Depth}
    \label{ShapeGT}
  \end{subfigure}
  \hspace{0em}
  \begin{subfigure}[b]{0.3\linewidth}
    \centering
    \includegraphics[width=\linewidth]{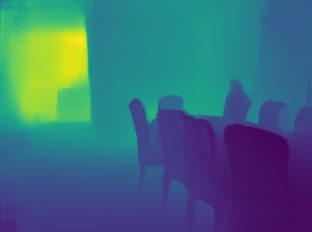}
    \caption{Estimated Depth}
    \label{ShapeOutput}
  \end{subfigure}
    \caption{Depth Estimation with a Shape Feature Input. The left and middle images are the original RGB image and corresponding ground truth depth map. The right is the estimated depth map from the model trained with shape maps.}
    \label{Shape}
\end{figure}

As we noted in Table \ref{allResults}, the shape comes across as the most dominant feature in these experiments, significantly outperforming other cues taken in isolation. We suggest this is because the data set contains indoor scenes of objects such as furniture with accurately extractable edges whose relative orientations and geometric forms can serve as powerful cues as seen in Figure \ref{Shape} (b and c).

The outcome aligns with the finding presented in \cite{hu2019visualization}, suggesting that CNNs are capable of deducing the depth map using merely a limited subset of pixels from the input image. This hypothesis aligns with human perceptual abilities, which allow for the extraction of approximate distance assessments from images that depict geometric shapes.

\subsubsection{Generalisation}

\begin{figure}[tb]
  \centering
\begin{subfigure}[b]{0.23\linewidth}
    \centering
    \includegraphics[width=\linewidth]{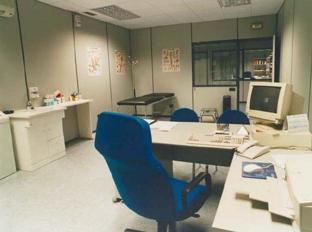}
  \end{subfigure}
  \hspace{-0.2em}
  \begin{subfigure}[b]{0.23\linewidth}
    \centering
    \includegraphics[width=\linewidth]{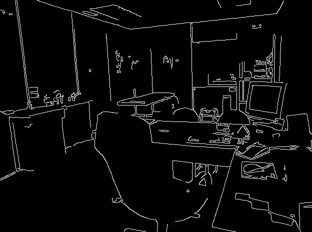}
  \end{subfigure}
  \hspace{-0.2em}
  \begin{subfigure}[b]{0.23\linewidth}
    \centering
    \includegraphics[width=\linewidth]{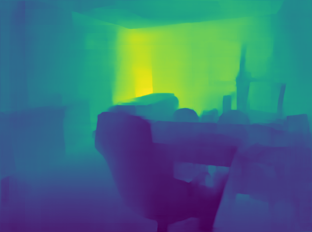}
  \end{subfigure}
  \hspace{-0.2em}
  \begin{subfigure}[b]{0.23\linewidth}
    \centering
    \includegraphics[width=\linewidth]{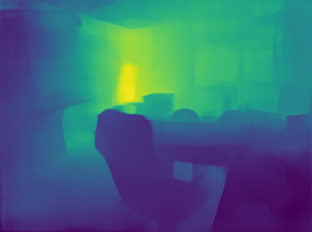}
  \end{subfigure}
    \centering
\begin{subfigure}[b]{0.23\linewidth}
    \centering
    \includegraphics[width=\linewidth]{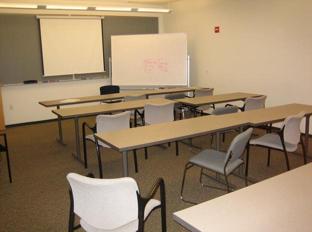}
  \end{subfigure}
  \hspace{-0.2em}
  \begin{subfigure}[b]{0.23\linewidth}
    \centering
    \includegraphics[width=\linewidth]{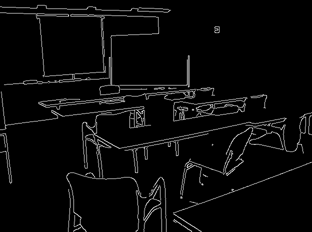}
  \end{subfigure}
  \hspace{-0.2em}
  \begin{subfigure}[b]{0.23\linewidth}
    \centering
    \includegraphics[width=\linewidth]{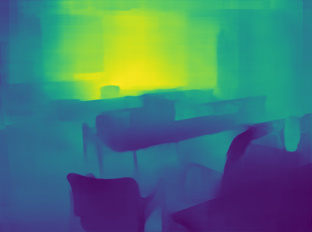}
  \end{subfigure}
  \hspace{-0.2em}
  \begin{subfigure}[b]{0.23\linewidth}
    \centering
    \includegraphics[width=\linewidth]{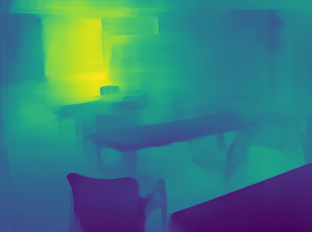}
  \end{subfigure}
    \centering
\begin{subfigure}[b]{0.23\linewidth}
    \centering
    \includegraphics[width=\linewidth]{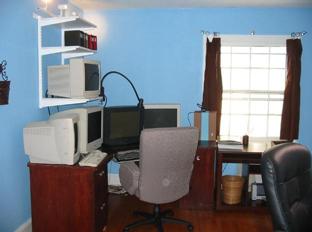}
  \end{subfigure}
  \hspace{-0.2em}
  \begin{subfigure}[b]{0.23\linewidth}
    \centering
    \includegraphics[width=\linewidth]{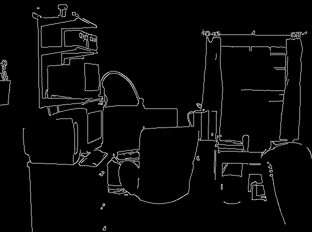}
  \end{subfigure}
  \hspace{-0.2em}
  \begin{subfigure}[b]{0.23\linewidth}
    \centering
    \includegraphics[width=\linewidth]{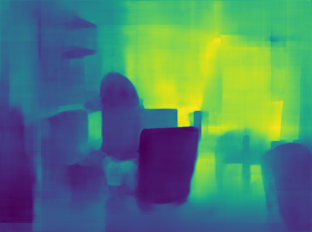}
  \end{subfigure}
  \hspace{-0.2em}
  \begin{subfigure}[b]{0.23\linewidth}
    \centering
    \includegraphics[width=\linewidth]{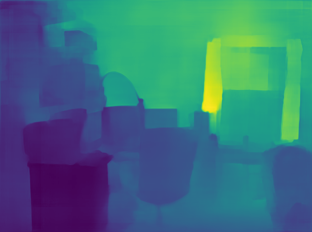}
  \end{subfigure}
    \centering
\begin{subfigure}[b]{0.23\linewidth}
    \centering
    \includegraphics[width=\linewidth]{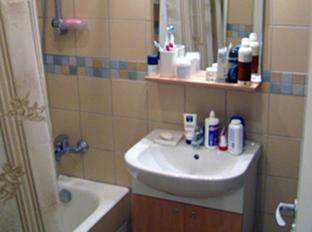}
  \end{subfigure}
  \hspace{-0.2em}
  \begin{subfigure}[b]{0.23\linewidth}
    \centering
    \includegraphics[width=\linewidth]{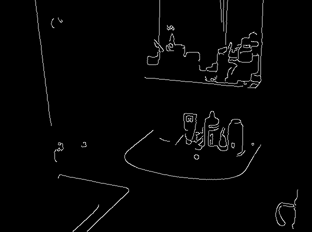}
  \end{subfigure}
  \hspace{-0.2em}
  \begin{subfigure}[b]{0.23\linewidth}
    \centering
    \includegraphics[width=\linewidth]{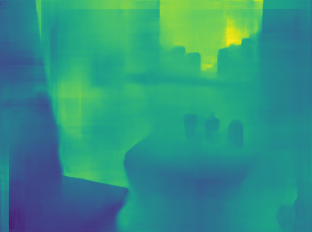}
  \end{subfigure}
  \hspace{-0.2em}
  \begin{subfigure}[b]{0.23\linewidth}
    \centering
    \includegraphics[width=\linewidth]{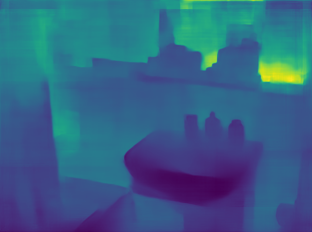}
  \end{subfigure}
      \caption{Performance of Shape ONLY model with New Indoor Scenes from other Domains. The left column displays original RGB scene images, the second column presents corresponding edge maps and the third column showcases the results generated by the pre-trained shape-input model. The right column exhibits the outcomes produced by the pre-trained original-RGB-image-input model.}
    \label{Shapes}
\end{figure}

In light of the fact that models using shape maps as input exhibit performance approximating that of models employing original RGB images as the input, we have assessed the generalisation capacity of shape models trained with shape maps on the NYU data set. We applied it to a diverse set of indoor environments from a different data set \cite{quattoni2009recognizing} that includes kitchens, bedrooms, bathrooms and various other scenes. The performance of the shape model is depicted in Figure \ref{Shapes}, illustrating its ability to predict depth maps even for scenes from a different domain, and the performance is similar to that of the original RGB model. However, shape maps, as input for depth estimation, still have their limitations. For instance, in the fourth-row images in Figure \ref{Shapes}, the sink only has partial edges, leading to poor depth prediction. Additional results are presented in the Appendix.

Shape maps require significantly less memory storage compared to original RGB images, while still providing comparable performance. In a similar vein, event cameras are designed to only detect rapid changes in pixel intensity \cite{rebecq2018esim,scheerlinck2020fast} which often occur at the edges of objects or where there is texture variation, which is similar to a shape map. Moreover, event cameras have previously been applied in the field of depth estimation \cite{gallego2020event}. Given these considerations, our research may offer supporting evidence for the application of event cameras in monocular depth estimation.

\subsubsection{Discussion}

Different types of input data have varying effects on the performance of depth estimation. Comparative analysis of diverse evaluation metrics clearly highlights the superior role of shape information in the depth estimation task. Colour, saturation and local texture collectively enhance the indoor scene depth estimation, although the influence of colour and saturation appears relatively circumscribed.

For phase-scrambled and local texture inputs, human vision finds it difficult to interpret images when their phase information is scrambled or only shuffled patches are present. In contrast, machines are adept at using these inputs to predict depth maps. Given that the models can output corresponding depth maps when employing these as inputs, their performance, albeit not optimal, is still noteworthy compared to the ability of humans.
\section{Limitations}
Throughout our study, we sought to isolate each feature we were evaluating. However, it is difficult to entirely isolate individual features. For instance, during the extraction of shape features, the edge detector might inadvertently capture some texture information.
\section{Conclusion}

In this work, we have sought to decouple and quantify the relative contributions of various depth cues in monocular depth estimation. Whereas good results have been demonstrated in the literature by the end-to-end training of deep neural network models to achieve this task, ours is the first attempt to understand the degree to which some known cues of depth contribute when taken in isolation. Our results show that, in a data set of indoor scenes, shape extracted by edge detection is relatively the most significant contributor, while other cues (colour, saturation and texture) also play a role. In achieving these conclusions, this work sought to carefully design feature extraction techniques that aimed to isolate a single feature from the other known ones, which is non-trivial. We speculate (and this is the subject of our current research) that, on different depth inference problems (e.g. outdoor scenes), the relative contributions of texture and saturation are likely to play a greater role. This kind of decomposition which we have extracted can serve to shift research more in the direction of understanding and explaining how powerful models, such as deep neural networks, work in scene understanding as opposed to simply offering estimation performance as black-box function approximators.

{
    \small

    \bibliographystyle{ieeenat_fullname}
}

\end{document}